\documentclass[11pt]{article}

\usepackage{times}
\usepackage[pdftex]{graphicx}
\usepackage{footmisc}
\usepackage{comment}
\usepackage{booktabs}
\usepackage{dcolumn}
\usepackage{hyperref}
\usepackage{url}
\usepackage{fullpage}
\usepackage{multirow}

\usepackage{subfigure}
\usepackage{color}

\usepackage{authblk}

\newcolumntype{d}[1]{D{.}{.}{#1} }

\begin{document}

\title{Cooperating with Machines}

\author[1]{Jacob W. Crandall\footnote{This is an incomplete pre-print of the following paper: Crandall {\em et al.}~Cooperating with Machines, {\em Nature Communications}, Vol 9, Article Number 233, 2018.  We encourage the reader to follow \href{https://www.nature.com/articles/s41467-017-02597-8}{\color{blue} this link} to view the published paper.}}
\author[2]{Mayada Oudah}
\author[2]{Tennom}
\author[2]{Fatimah Ishowo-Oloko}
\author[3,4]{Sherief~Abdallah}
\author[5]{\\Jean-Fran\c{c}ois Bonnefon}
\author[6]{Manuel Cebrian}
\author[7]{Azim Shariff}
\author[1]{Michael A. Goodrich}
\author[8]{Iyad~Rahwa${\rm n}^*$}
\affil[1]{Computer Science Department, Brigham Young University, Provo, UT 84602, USA}
\affil[2]{Masdar Institute of Science and Technology, Abu Dhabi, UAE}
\affil[3]{British University in Dubai, Dubai, UAE}
\affil[4]{School of Informatics, University of Edinburgh, Edinburgh EH8 9AB, UK}
\affil[5]{Toulouse School of Economics, Center for Research in Management, Centre National de la Recherche Scientifique, Institute for Advanced Study in Toulouse, University of Toulouse Capitole, Toulouse, France}
%\affil[5]{Centre National de la Recherche Scientifique, Toulouse, France}
\affil[6]{Data61, Commonwealth Scientific and Industrial Research Organization, Clayton, Victoria 3168, Australia}
\affil[7]{Department of Psychology and Social Behavior, University of California, Irvine, CA 92697, USA}
\affil[8]{The Media Lab, Massachusetts Institute of Technology, Cambridge, MA 02139, USA}

\date{}
\maketitle

\begin{abstract}
Since Alan Turing envisioned Artificial Intelligence (AI)~\cite{turing1950computing}, a major driving force behind technical progress has been competition with human cognition. Historical milestones have been frequently associated with computers matching or outperforming humans in difficult cognitive tasks (e.g. face recognition~\cite{toole2007face}, personality classification~\cite{youyou2015computer}, driving cars~\cite{Montemerlo2008}, or playing video games \cite{mnih2015human}), or defeating humans in strategic zero-sum encounters (e.g. Chess~\cite{campbell2002deep}, Checkers~\cite{schaeffer2007checkers}, Jeopardy!~\cite{ferrucci2010building}, Poker~\cite{bowling2015heads}, or Go~\cite{Go}). In contrast, less attention has been given to developing autonomous machines that establish mutually cooperative relationships with people who may not share the machine's preferences.  A main challenge has been that human cooperation does not require sheer computational power, but rather relies on intuition~\cite{rand2014social}, cultural norms~\cite{boyd2009culture}, emotions and signals~\cite{Frank1988,Skyrms2003,DavidSally,Balliet}, and pre-evolved dispositions toward cooperation~\cite{peysakhovich2014humans}.  Here, we combine a state-of-the-art machine-learning algorithm with novel mechanisms for generating and acting on signals to produce a new learning algorithm that cooperates with people and other machines at levels that rival human cooperation in a variety of two-player repeated stochastic games. This is the first algorithm that is capable of learning to cooperate with people within short timescales in scenarios previously unanticipated by algorithm designers.  This is achieved without complex opponent modeling or higher-order theories of mind, thus showing that flexible, fast, and general human-machine cooperation is computationally achievable using a non-trivial, but ultimately simple, set of algorithmic mechanisms.
\end{abstract}

%These common-sense mechanisms can be difficult to effectively encode in autonomous machines.  

\pagenumbering{gobble}

\newpage

\section{Introduction}

The emergence of driverless cars, autonomous trading algorithms, and autonomous drone technologies highlight a larger trend in which artificial intelligence (AI) is enabling machines to autonomously carry out complex tasks on behalf of their human stakeholders.  To effectively represent their stakeholders in many tasks, these autonomous machines must repeatedly interact with other people and machines that do not fully share the same goals and preferences.  While the majority of AI milestones have focused on developing human-level wherewithal to compete with people~\cite{campbell2002deep,schaeffer2007checkers,ferrucci2010building,bowling2015heads,Go}, most scenarios in which AI must interact with people and other machines are not zero-sum interactions.  As such, AI must also have the ability to cooperate, even in the midst of conflicting interests and threats of being exploited.  Our goal is to understand how to build AI algorithms that cooperate with people and other machines at levels that rival human cooperation in arbitrary two-player repeated interactions.

Algorithms capable of forming cooperative relationships with people and other machines in arbitrary scenarios are not easy to come by.  A successful algorithm should possess several properties.  First, it must not be domain-specific -- it must have superior performance in a wide variety of scenarios ({\em generality}).  Second, the algorithm must learn to establish effective relationships with people and machines without prior knowledge of associates' behaviors ({\em flexibility}).  To do this, it must be able to deter potentially exploitative behavior from its partner and, when beneficial, determine how to elicit cooperation from a (potentially distrustful) partner who might be disinclined to cooperate.  Third, when associating with people, the algorithm must learn effective behavior within very short timescales -- i.e., within only a few rounds of interaction ({\em learning speed}).  These requirements create many technical challenges (see SI.A.2), the sum of which often causes AI algorithms to fail to cooperate, even when doing so would be beneficial in the long run.

In addition to these computational challenges, human-AI cooperation is difficult due to differences in the way that humans and machines reason.  While AI relies on computationally intensive search and random exploration to generate strategic behavior, human cooperation appears to rely on intuition~\cite{rand2014social}, cultural norms~\cite{boyd2009culture}, emotions and signals~\cite{Frank1988,Skyrms2003}, and pre-evolved dispositions toward cooperation~\cite{peysakhovich2014humans}.   In particular, cheap talk (i.e., costless signals) is important to human cooperation in repeated interactions~\cite{DavidSally,Balliet}, as it helps people coordinate quickly on desirable equilibrium and create shared representations~\cite{klein2005common, dautenhahn2007socially, breazeal2003toward, kamar2013modeling}.  As such, in addition to generating effective strategic behavior, a fourth important property of a successful algorithm designed for repeated interactions is the ability to {\em communicate} effectively at levels conducive to human understanding.

%we consider that AI algorithms must generate and respond to costless signals at levels that are conducive to human understanding.

\section{Results}

The primary contribution of this work is the development and analysis of a new learning system that couples a state-of-the-art machine-learning algorithm with novel mechanisms for generating and responding to signals.  Via extensive simulations and user studies, we show that this learning system learns to establish and maintain effective relationships with people and other machines in a wide-variety of repeated interactions at levels that rival human cooperation.  In so doing, we also investigate the algorithmic mechanisms that are responsible for its success.

%Our goal is to understand how to build AI algorithms that have the ability to cooperate with both people and other machines in long-term repeated interactions.  To do so, we conducted a vast number of simulations and user studies.  In the simulations, we compared the performance of 25 distinct algorithms, and evaluated their abilities to interact with each other.  Then, via a series of three user studies, we evaluated the ability of selected algorithms (including a new algorithm developed in this work) to forge successful relationships with people over repeated interactions.

\subsection{An Algorithm that Cooperates with People and Other Machines}

Over the last several decades, algorithms for generating strategic behavior in repeated games have been developed in many disciplines, including economics, evolutionary biology, and the AI and machine-learning communities.  To begin to evaluate the ability of these algorithms to forge successful cooperative relationships, we selected and evaluated 25 representative algorithms from these fields, including classical algorithms such as (generalized) generous tit-for-tat (i.e., {\sc Godfather}) and win-stay-lose-shift (WSLS)~\cite{Nowak1993}, evolutionarily evolved memory-one and memory-two stochastic strategies~\cite{Iliopoulous2010}, machine-learning algorithms (including reinforcement learning), belief-based algorithms~\cite{FictitiousPlay}, and expert algorithms~\cite{exextrade,CrandallJAIR2014}.  Via extensive simulations, we compared these algorithms with respect to six different performance metrics (see SI.B.2) across the periodic table of 2x2 games~\cite{robinsongoforth_book} (see Methods and SI.A.3).  

The results of this evaluation, which are overviewed in Methods (see Figure~\ref{fig:perfSummary} in particular) and are described in detailed in SI.B, demonstrate the difficulty of developing algorithms that can forge effective long-term relationships in many different scenarios.  The results show that only S++~\cite{CrandallJAIR2014} was a top-performing algorithm across all metrics at all game lengths when associating with other algorithms.  However, despite its fast learning speeds and its success in interacting with other machines in many different scenarios, S++ does not, in its current form, consistently forge cooperate relationships with people (SI.D and SI.E.1-2), though it does cooperate with people as frequently as people cooperate with each other in the same studies.  Thus, none of these existing algorithms establishes effective long-term relationships with both people and machines.

We hypothesized that S++'s inability to consistently learn to cooperate with people appears to be tied to its inability to generate and respond to costless signals.  Humans are known for their ability to effectively coordinate on cooperative equilibria using costless signals called cheap talk~\cite{DavidSally,Balliet}.  However, while signaling comes naturally to humans, the same cannot be said of sophisticated AI algorithms, such as machine-learning algorithms.  To be useful, costless signals should be connected with behavioral processes.   Unfortunately, most machine-learning algorithms have low-level internal representations that are often not easily expressed in terms of high-level behavior, especially in arbitrary scenarios.  As such, it is not obvious how these algorithms can be used to generate and respond to costless signals at levels that people understand.

\begin{figure}[!p]
\begin{center}
\includegraphics[width=6.0in]{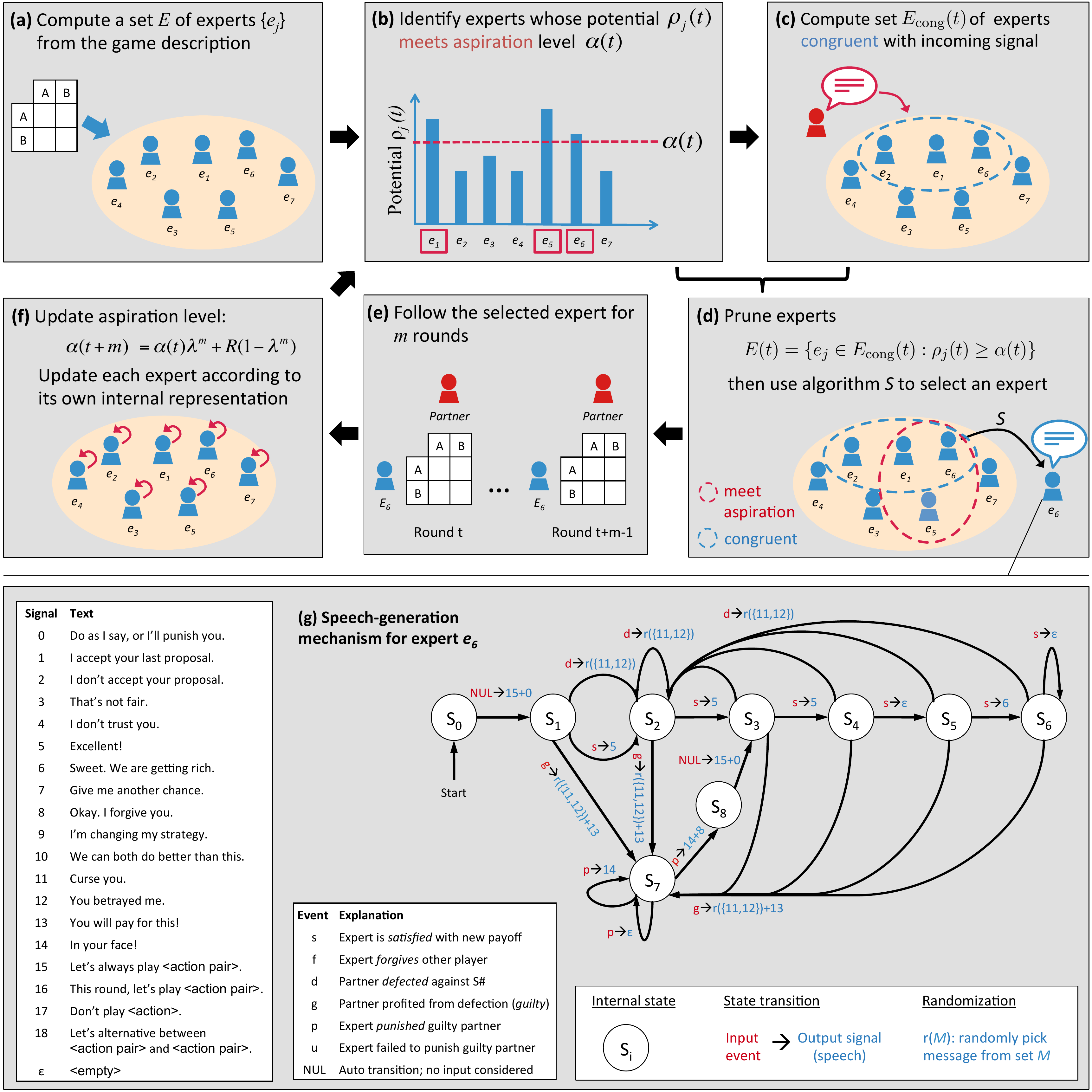} \vspace{-.1in}
\end{center}
\caption{\small An overview of S\#, an algorithm that interweaves signaling capabilities into S++~\cite{CrandallJAIR2014}.  {\bf (a)}~Prior to beginning the game, S\# uses the description of the game to compute a set $E$ of expert strategies.  Each expert encodes a strategy or learning algorithm that defines behavior over all game states.  {\bf (b)}~S\# computes the potential, or highest expected utility, of each expert in $E$.  The potentials are then compared to an aspiration level $\alpha(t)$, which encodes the average per-round payoff that the algorithm believes is achievable, to determine a set of experts that could potentially meet the agent's aspirations.  {\bf (c)}~S\# determines which experts carry out plans that are congruent with its partner's last proposed plan.  {\bf (d)}~S\# selects an expert (using algorithm S~\cite{Karandikar,StimpsonIJCAI}) from among those experts that both potentially meet its aspirations (step b) and are congruent with its partner's latest proposal (step c).  If $E(t)$ is empty, S\# selects its expert from among the set of experts that meet its aspiration level (step b).  The currently selected expert generates signals based on its game-generic state machine (bottom).  Given the current state of the expert and game events, the expert produces speech from a predetermined list of speech acts.  {\bf (e)}~The machine follows the strategy dictated by the selected expert for $m$ rounds of the repeated game.  {\bf (f)}~The machine updates its aspiration level based on the average reward $R$ it has received over the last $m$ rounds of the game.  The experts are also updated according to their own internal representations.  The algorithm then returns to step b.  The process repeats for the duration of the repeated game.  Details are given in SI.C.  Note that S++ is identical to S\# except that S++ (1)~replaces step c with $E_{\rm cong}(t) = E$, and (2)~does not generate speech acts.}
\label{fig:Ssharp} 
\end{figure}

Fortuitously, unlike typical machine-learning algorithms, the internal structure of S++ provides a clear, high-level representation of the algorithm's dynamic strategy that can be described in terms of the dynamics of the underlying experts.  Since each expert encodes a high-level philosophy, S++ could potentially be used to generate signals (i.e., cheap talk) that describe its intentionality.  Speech acts from its partner can also be compared to its experts' philosophies to improve its expert-selection mechanism.  In this way, S++ can be augmented with a communication framework that gives it the ability to generate and respond to cheap talk.  The resulting new algorithm, dubbed S\# (pronounced `S sharp'), is depicted in Figure~\ref{fig:Ssharp} (see Methods and SI.C for details about the algorithm).  

%Humans are known for their ability to effectively coordinate on cooperative equilibria using costless signals called cheap talk~\cite{DavidSally,Balliet}.  Thus, S++'s inability to consistently learn to cooperate with people appears to be tied to its inability to generate and respond to these costless signals.  While signaling comes naturally to humans, the same cannot be said of machine-learning algorithms.  To be useful, costless signals should be connected with behavioral processes.   Unfortunately, most machine-learning algorithms have low-level internal representations that are often not easily expressed in terms of high-level behavior, especially in arbitrary scenarios.  As such, it is not obvious how these algorithms can be used to generate and respond to costless signals at levels that people understand.  However, unlike typical machine-learning algorithms, the internal structure of S++ provides a clear, high-level representation of the algorithm's dynamic strategy that can be described in terms of the dynamics of the underlying experts.  Since each expert encodes a high-level philosophy, we can use S++ to generate signals (i.e., cheap talk) that describe its intentionality.  Speech acts from its partner can also be compared to its experts' philosophies to improve its expert-selection mechanism.

We conducted a series of three user studies (see SI.D--F for details) involving 220 participants, who played in a total of 472 games, to determine the ability of S\# to forge cooperative relationships with people.  Representative results are found in the final (culminating) study, in which participants played three representative repeated games (drawn from distinct payoff families; see SI.A.3) via a computer interface that hid the identity of their partner.  In some conditions, players could engage in cheap talk by sending messages at the beginning of each round via the computer interface.  Consistent with prior work investigating cheap talk in repeated games~\cite{Balliet}, messages were limited to the predetermined speech acts available to S\#.  %Interactions between players when cheap talk was permitted are logged in SI.G.

\begin{figure}[!p]
\begin{center}
\includegraphics[width=6.45in]{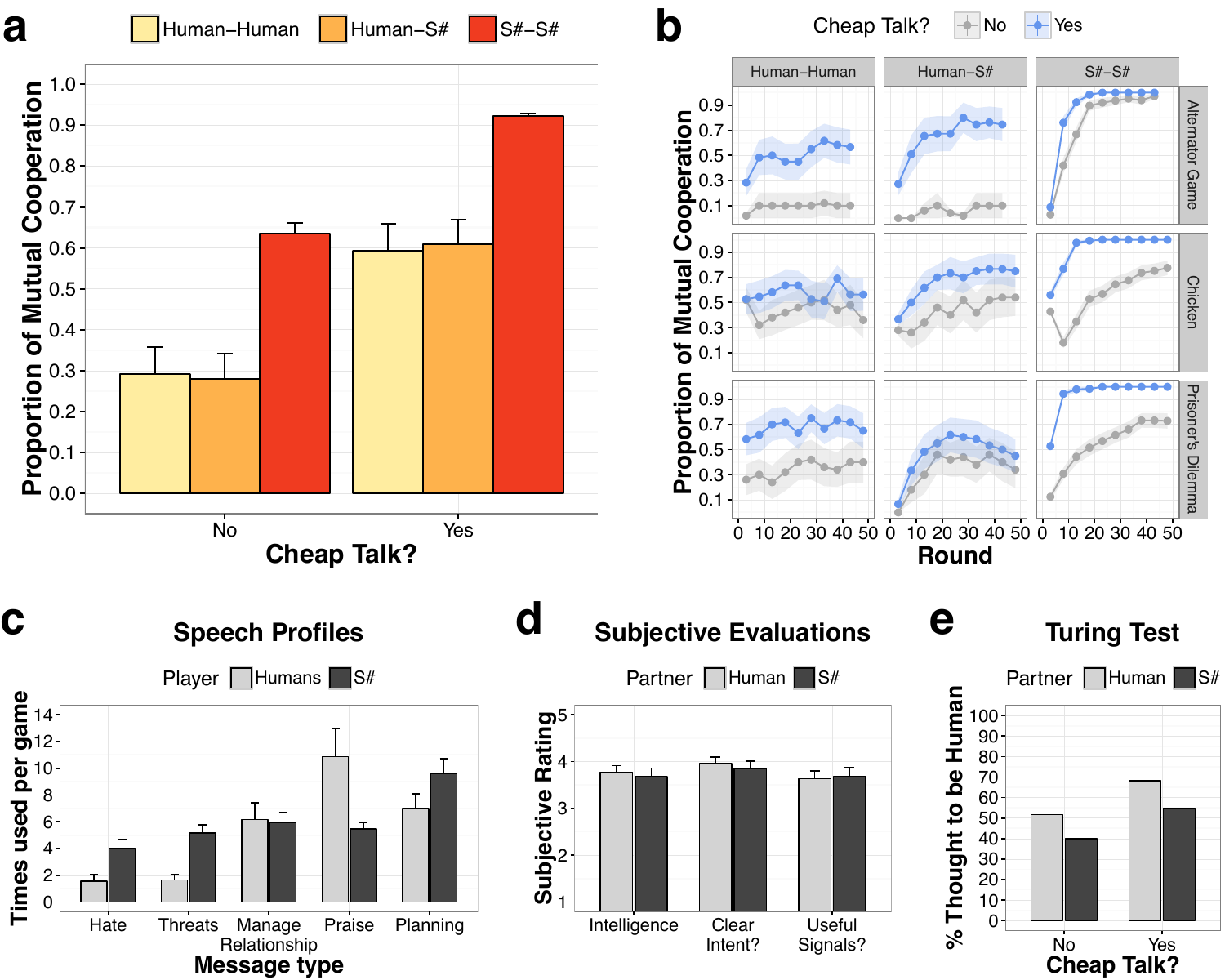}
\end{center} \vspace{-.15in}
\caption{Results of the culminating user study in which 66 volunteer participants (people) were paired with each other and S\# in three representative games (Chicken, Alternator Game, and Prisoner's Dilemma).  S\# is identical to S++ when cheap talk is not permitted.  Bars and lines show average values over all trials, while error bars and ribbons show the standard error of the mean.  Full details related to sample size and statistical tests are provided in SI.F.  {\bf (a)}~The average proportion of mutual cooperation across all three games under conditions in which cheap talk between players was either permitted or not permitted.  {\bf (b)}~The average proportion of mutual cooperation over time in each game in each pairing and condition.  {\bf (c)} The average number of times that Humans and S\# used messages of each type over the course of an interaction when paired with people across all games.  For simplicity, the 19 speech acts were grouped into five categories (see SI.F.1.3).  S\# tended to use more negative speech acts (labeled {\em Hate} and {\em Threats}), while people tended to use more positive speech acts ({\em praise}). {\bf (d)} Results of three post-experiment questions for subjects that experienced the condition in which cheap talk was permitted.  Participants rated (1)~the intelligence of their partner, (2)~the clarity of their partner's intentions, and (3)~the usefulness of the communication between them and their partner.  Answers were given on a 5-point Likert Scale.  Specific questions and scales are provided in SI.F. {\bf (e)}~The percentage of time that human participants and S\# were thought to be human by their partner when cheap talk was both permitted and not permitted.}
\label{fig:mainresults} 
\end{figure}

The proportion of mutual cooperation achieved by Human-Human, Human-S\#, and S\#-S\# pairings are shown in Figures~\ref{fig:mainresults}a-b.  When cheap talk was not permitted, Human-Human and Human-S\# pairings did not frequently result in cooperative relationships.  However, across all three games, the presence of cheap talk doubled the proportion of mutual cooperation experienced by these two pairings.  While S\#'s speech profile was distinct from that of humans (Figure~\ref{fig:mainresults}c), subjective, post-interaction assessments indicate that S\# used cheap talk to promote cooperation as effectively as people (Figure~\ref{fig:mainresults}d).   In fact, many participants were unable to distinguish S\# from a human player (Figure~\ref{fig:mainresults}e).  Together, these results illustrate that, across the games studied, the combined behavioral and signaling strategies of S\# were as effective as those of human players.

\subsection{Distinguishing Algorithmic Mechanisms}

\begin{figure}[!p]
\begin{center}
\includegraphics[width=6.0in]{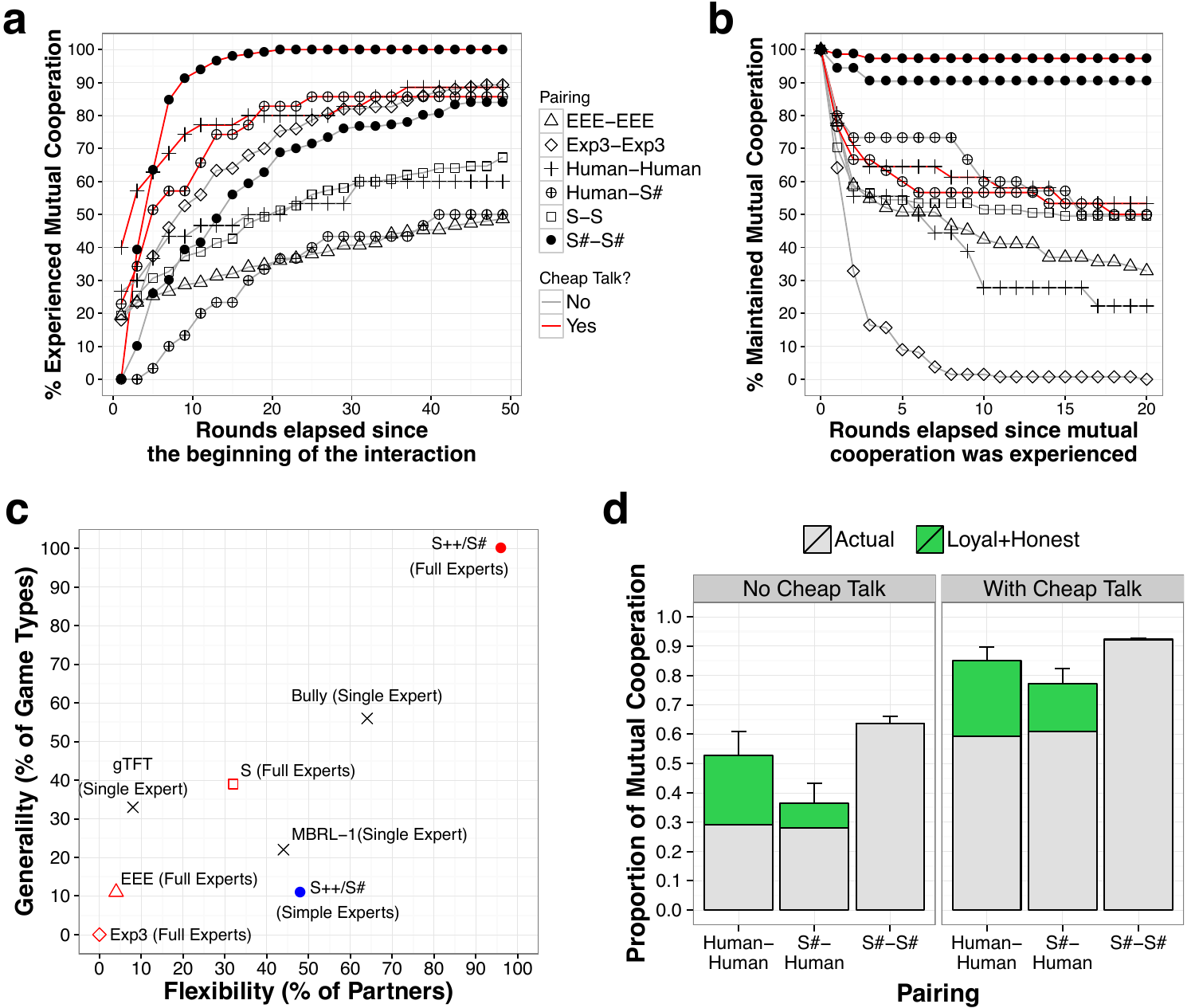}
\end{center}
\caption{{\bf (a)} Empirically generated cumulative distribution functions for the number of rounds required for pairings to experience two consecutive rounds of mutual cooperation across three different repeated games (Chicken, Alternator Game, and Prisoner's Dilemma).  Per-game results are provided in SI.F.  For machine-machine pairings, the results are obtained from 50 trials conducted in each game, whereas pairings with humans use results from a total of 36 different pairings each.  {\bf (b)} The percentage of partnerships for each pairing that did not deviate from mutual cooperation once the players experienced two consecutive rounds of mutual cooperation across the same three repeated games. {\bf (c)} The percentage of game types (payoff family $\times$ game length) and partners (25 different algorithms) against which various algorithms were ranked in the top 2 (among the 25 different algorithms considered) with respect to payoffs received.  See SI.B.5 for details.  {\bf (d)}~The estimated proportion of rounds that would have resulted in mutual cooperation had all human players followed S\#'s learned behavioral and signaling strategies of (1)~not deviating from cooperative behavior when mutual cooperation was established (i.e., {\em Loyal}) and (2)~following through with verbal commitments (i.e., {\em Honest}).  See SI.F.4 for details.  Error bars show the standard error of the mean.  Had all human participants been loyal and honest, these results indicate that there would have been little difference between Human-Human and S\#-S\# pairings.  }\vspace{.3in}
\label{fig:unpack}
\end{figure}

Why is S\# so successful in forging cooperative relationships with both people and other algorithms?  Are its algorithmic mechanisms fundamentally different from those of other algorithms for repeated games?  We have identified three algorithmic mechanisms responsible for S\#'s success.  Clearly, Figure~\ref{fig:mainresults} demonstrates that the first of these mechanisms is S\#'s ability to generate and respond to relevant signals people can understand, a trait not present in previous learning algorithms designed for repeated interactions. These signaling capabilities expand S\#'s flexibility in that they also allow S\# to more consistently forge cooperative relationships with people.  Without this capabilities, it does not consistently do so.  Figure~\ref{fig:unpack}a demonstrates one simple reason that this mechanism is so important: cheap talk helps both S\# and humans to more quickly experience mutual cooperation with their partners.

Second, our implementation of S\# uses a rich set of experts that includes a variety of equilibrium strategies and even a simple learning algorithm (see SI.C.1).  While none of these individual experts has an overly complex representation (e.g., no expert remembers the full history of play), these experts are more sophisticated than those traditionally considered (though not explicitly excluded) in the discussion of expert algorithms~\cite{auer95gambling,GIGA-WoLF,FreundSchapire95}.  This more sophisticated set of experts permits S\# to adapt to a variety of partners and game types, whereas algorithms that rely on a single strategy or a less sophisticated set of experts are only successful in particular kinds of games played with particular partners~\cite{Marstaller2013} (Figure~\ref{fig:unpack}c).  Thus, in general, simplifying S\# by removing experts from this set will tend to limit the algorithm's flexibility and generality, though doing so will not always negatively impact its performance when paired with particular associates in particular games.

Finally, S\#'s somewhat non-conventional expert-selection mechanism (see Eq.~\ref{eq:theset}) is central to its success.  While techniques such as $\varepsilon$-greedy exploration (e.g., EEE) and regret-matching (e,g., Exp3) have permeated algorithm development in the AI community, S\# instead uses an expert-selection mechanism closely aligned with recognition-primed decision making \cite{Klein1989}.  Given the same full, rich set of experts, more traditional expert-selection mechanisms establish effective relationships in far fewer scenarios than S\# (Figure~\ref{fig:unpack}c).  Figures~\ref{fig:unpack}a-b provide insights into why this is so.  Compared to the other expert-selection mechanisms, S\# has a greater combined ability to quickly establish a cooperative relationship with its partner (Figure~\ref{fig:unpack}a) and then to maintain it (Figure~\ref{fig:unpack}b), a condition brought about by S\#'s tendency to not deviate from cooperation after mutual cooperation has been established (i.e., loyalty).

The loyalty brought about by S\#'s expert-selection mechanism helps explain why S\#-S\# pairings substantially outperformed Human-Human pairings in our study (Figure~\ref{fig:mainresults}a-b).  S\#'s superior performance can be attributed to two human tendencies.  First, while S\# did not typically deviate from cooperation after successive rounds of mutual cooperation (Figure~\ref{fig:unpack}b), many human players did.  Almost universally, such deviations led to reduced payoffs to the deviator.  Second, a sizable portion of our participants failed to keep some of their verbal commitments.  On the other hand, since S\#'s verbal commitments are derived from its intended future behavior, it typically carries out the plans it proposes.  Had participants followed S\#'s strategy in these two regards, Human-Human pairings would have performed nearly as well, on average, as S\#-S\# pairings (Figure~\ref{fig:unpack}d -- see SI.F.4 for details).

\subsection{Repeated Stochastic Games}

\begin{figure}
\begin{center}
\includegraphics[width=6.0in]{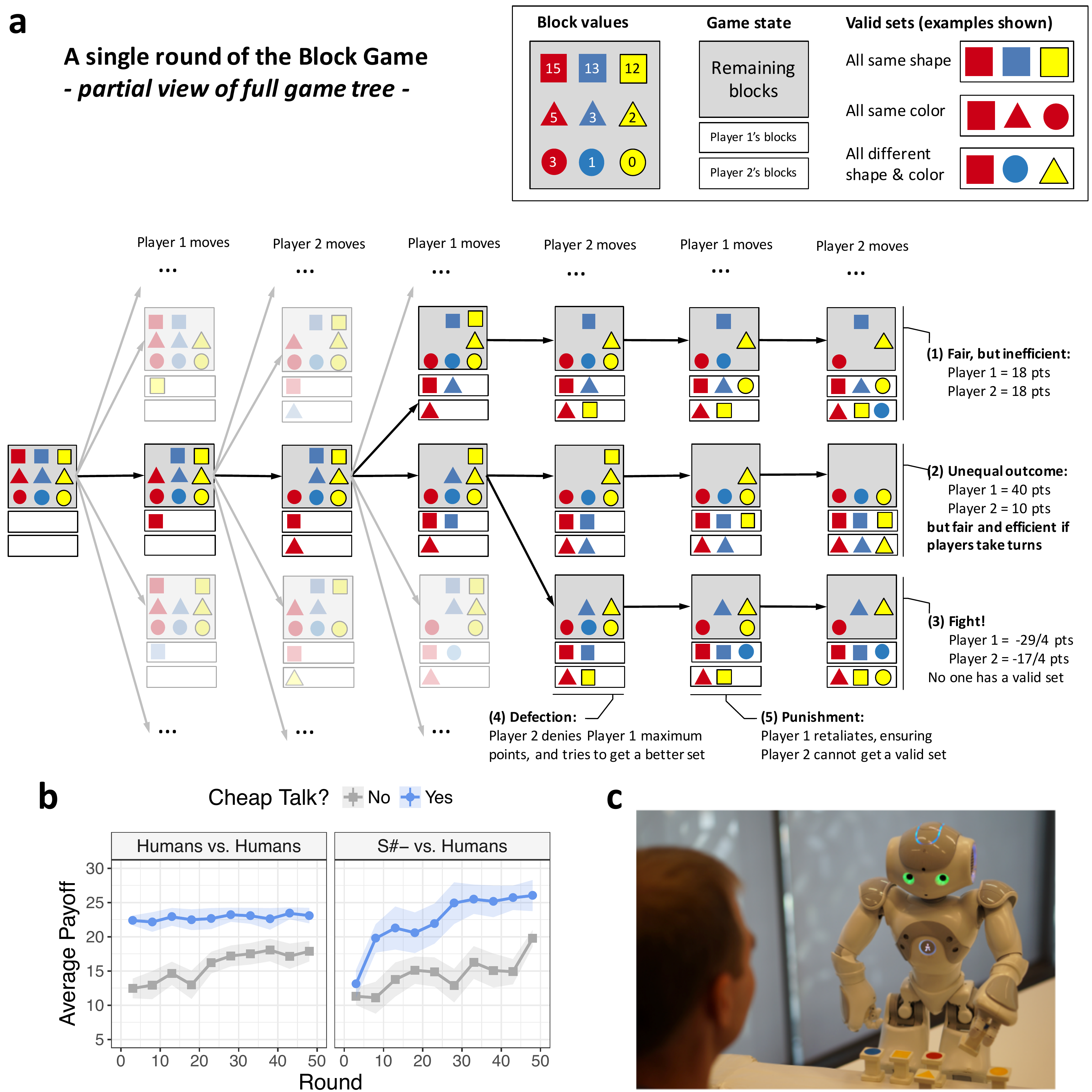} \vspace{-.2in}
\end{center}
\caption{{\small In addition to evaluating algorithms in normal-form games, we also evaluated people and algorithms in repeated stochastic games (including extensive-form games).  Results are provided in SI.E.  {\bf (a)} An extensive-form game in which two players share a nine-piece block set.  The two players take turns selecting blocks from the set until each has three blocks.  The goal of each player is to get a {\em valid} set of blocks with the highest value possible, where the value of a set is determined by the sum of the numbers on the blocks.  Invalid sets receive negative points.  (1)~A fair, but inefficient outcome in which both players receive 18 points.  (2)~An unequal outcome in which one player receives 40 points, while the other player receives just 10 points.   However, when the players take turns getting the higher payoff (selecting all the squares), this is the Nash bargaining solution of the game, producing an average payoff of 25 to both players.  (3)~An outcome in which neither player obtains a valid set, and hence both players lose points.  (4)~This particular negative outcome is brought about when player 2 defects against player 1 by taking the block that player 1 needs to complete its (most-valuable) set.  (5)~Player 1 then retaliates to ensure that player 2 does not get a valid set either.  {\bf (b)}~Average payoffs obtained by people and S\#- (an early version of S\# that generates, but does not respond to, cheap talk) when associating with people in the extensive-form game depicted in a.  As in normal-form games, S\#- successfully uses cheap talk to consistently forge cooperative relationships with people in this repeated stochastic game.  For more details see SI.E.  {\bf (c)}~We also implemented S\#- on a Nao robot to play the Block Game with people.}}
\label{fig:blocks}
\end{figure}

The previous results were demonstrated for normal-form games.  However, S++ also learns effectively in repeated stochastic games~\cite{crandallIJCAI2015}, which are more complex scenarios in which a round consists of a sequence of moves by both players.  In these games, S++ is distinguished, again, by its ability to adapt to many different machine associates in a variety of different scenarios~\cite{crandallIJCAI2015}.  As in normal-form games, S++ can be augmented with cheap talk to form S\#.  While S++ does not consistently forge effective relationships with people in these more complex scenarios, our results show that S\# does.  Representative results are shown in Figure~\ref{fig:blocks}, which considers a turning-taking scenario in which two players must learn how to share a set of blocks.  Like people, S\# uses cheap talk to substantially increase its payoffs when associating with other people in this game (Figure~\ref{fig:blocks}b).  These results mirror those we observe in normal-form games (compare Figures~\ref{fig:blocks}b and~\ref{fig:mainresults}b).  See SI.E for additional details and results.%While S++ (S\# without the ability to engage in cheap talk) fails to establish cooperative relationships with people, while S\# consistently does so.  Details and additional results are provided in SI.E.

%It does so by learning to cooperate with associates that are inclined to cooperate when cooperation is beneficial, and effectively avoiding being exploited otherwise.  

%and S\# obtains similar results in more complex scenarios (such as repeated stochastic games) in which a round consists of a sequence of moves by both players~\cite{crandallIJCAI2015,Oudah2015}.  Representative results are shown in Figure~\ref{fig:blocks}, which details a block-sharing (turning-taking) game between two players.  Despite the added complexity of the game, like people, S\# uses cheap talk to great improve its payoffs when associating with other people (Figure~\ref{fig:blocks}b).  Details and additional results are provided in SI.E.

\section{Discussion}

Our studies of human-S\# partnerships were limited to five repeated games, selected carefully to represent different classes of games from the periodic table of games (see SI.A.3).  Though future work should address more scenarios, S\#'s success in establishing cooperative relationships with people in these representative games, along with its consistently high performance across all classes of 2x2 games and various repeated stochastic games~\cite{crandallIJCAI2015} when associating with other algorithms, gives us some confidence that these results will generalize to other scenarios.

Since Alan Turing envisioned Artificial Intelligence, major milestones have focused on defeating humans in zero-sum encounters~\cite{campbell2002deep,schaeffer2007checkers,ferrucci2010building,bowling2015heads,Go}.  However, in many scenarios, successful machines must cooperate with, rather than compete against, humans and other machines, even in the midst of conflicting interests and threats of being exploited.  Our work demonstrates how autonomous machines can learn to establish cooperative relationships with people and other machines in repeated interactions.  We showed that human-machine and machine-machine cooperation is achievable using a non-trivial, but ultimately simple, set of algorithmic mechanisms.  These mechanisms include computing a variety of \emph{expert} strategies optimized for various scenarios, a particular meta-strategy for a particular meta-strategy for \emph{selecting experts to follow}, and the ability to generate and respond to simple \emph{signals}. We hope that this first extensive demonstration of human cooperation with autonomous machines in repeated games will spur significant further research that will ensure that autonomous machines, designed to carry out human endeavors, will cooperate with humanity.

\section{Methods}

Detailed methods and analysis are provided in the SI.  In this section, we overview three different aspects of these methods and analysis: the benchmark of games used to compare algorithms and people, results from our comparison of AI algorithms, and a description of S\#.

\begin{figure}[!p]
\begin{center}
\vspace{-.25in}
\includegraphics[width=6.1in]{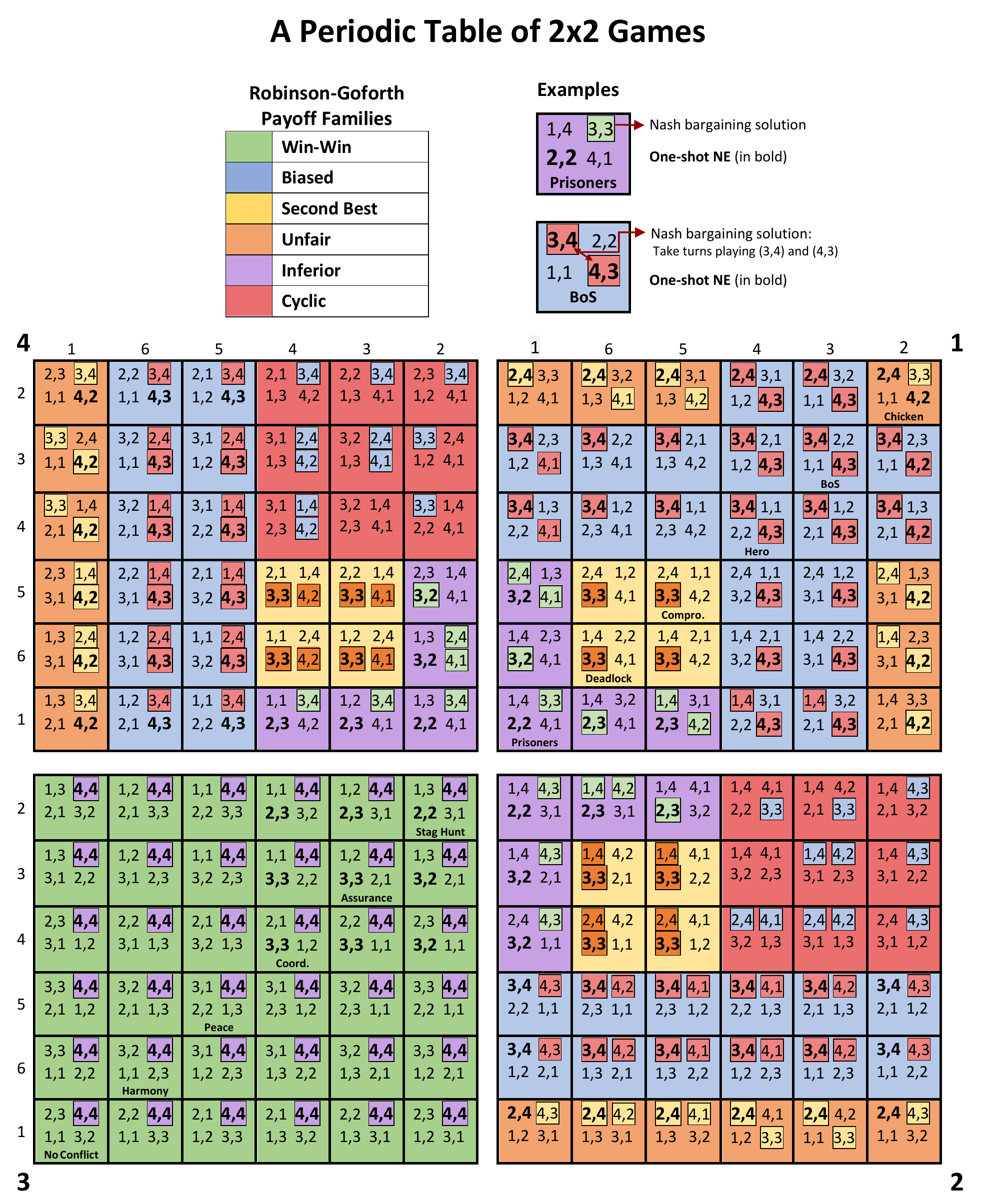} \vspace{-.2in}
\end{center}
\caption{We compared algorithms across the periodic table of 2x2 games based on the topology of Robinson and Goforth~\cite{robinsongoforth_book} for scenarios in which the players exhibit a strict ordinal preference ordering over the four game outcomes (specified by the values 1, 2, 3, and 4).  For each game, the pure-strategy one-shot Nash equilibria (NEs) are given in bold-face type.  The solutions played in the Nash bargaining solution (NBS~\cite{NashBargain} -- i.e., the mutually cooperative solution) given the payoff values 1, 2, 3, and 4 are also highlighted, though the frequency at which each solution is played is not specified.  Note that since the NBS depends on the actual payoffs and not just the preference ordering, other NBSs are possible for each game structure. The figure is adapted from the graphic developed by Bruns~\cite{bruns2010}.}
\label{fig:periodic_table}
\end{figure}

\subsection{Benchmark Games for Studying Cooperation}
As with all historical grand challenges in AI, it is important to identify a class of benchmark problems to compare the performance of different algorithms. When it comes to human cooperation, a fundamental benchmark has been $2\times 2$, general-sum, repeated games \cite{rapoport1967taxonomy}. This class of games has been a workhorse for decades in the fields of behavioral economics \cite{camerer2010behavioral}, mathematical biology \cite{hofbauer1998evolutionary}, psychology \cite{rand2013human}, sociology \cite{kollock1998social}, computer science~\cite{LittmanRepeatedNash_2}, and political science \cite{Axelrod1984}. These fields have revealed many aspects of human cooperative behavior through canonical games, such as the Prisoner's Dilemmas, Chicken, Battle of the Sexes, and the Stag Hunt. Such games, therefore, provide a well-established, extensively studied, and widely understood benchmark for studying the capabilities of machines to develop cooperative relationships.

The periodic table of $2\times 2$ games (Figure~\ref{fig:periodic_table}; see SI.A.3;~\cite{rapoport1967taxonomy,rapoport2,theoryofmoves,robinsongoforth_book,bruns2010}) identifies and categorizes 144 unique game structures that present many unique scenarios in which machines may need to cooperate.  We use this set of game structures as a benchmark against which to compare the abilities of algorithms to cooperate. Successful algorithms should be able to forge successful relationships with both people and machines across all of these repeated games. In particular, we can use these games to quantify the abilities of various state-of-the-art machine learning algorithms to satisfy the aforementioned properties: generality across games, flexibility across opponent types (including humans), and speed of learning.

Like the majority of work in repeated interactions, we focus on two-player normal-form games to more easily understand how machines can forge cooperative relationships with people.  Nevertheless, we are interested in algorithms that can also be used in more complex interactions, including the more general case of repeated (two-player) stochastic games (see, for example, Figure~\ref{fig:blocks}).  Studies evaluating the ability of S\# to forge cooperative relationships with people in repeated stochastic games have yielded similar results to those we report for two-player normal-form games (e.g., Figure~\ref{fig:blocks}b).  These studies are described in SI.E.

% (see, for example, Figure~\ref{fig:blocks})

\subsection{Interacting with Other Machines: AI Algorithms for Repeated Interactions}

%Over the last several decades, algorithms for generating strategic behavior in repeated games have been developed in many disciplines, including economics, evolutionary biology, and the AI and machine-learning communities.  With the goal of identifying successful algorithmic mechanisms for playing arbitrary repeated games, we selected and evaluated 25 representative algorithms from these fields (see Figure~\ref{fig:perfSummary}a).  These algorithms include classical algorithms such as (generalized) generous tit-for-tat (i.e., {\sc Godfather}) and win-stay-lose-shift (WSLS)~\cite{Nowak1993}, evolutionarily evolved memory-one and memory-two stochastic strategies~\cite{Iliopoulous2010}, machine-learning algorithms (including reinforcement learning), belief-based algorithms~\cite{FictitiousPlay}, and expert algorithms~\cite{exextrade,CrandallJAIR2014}.  

\begin{figure}[!p]
\begin{center}
\subfigure[Rankings of algorithms across six different metrics at three different game lengths]{
{\scriptsize
\begin{tabular}{l|cccccc|c} \hline

\multirow{2}{*}{\bf Algorithm} & {\bf Round-Robin} & {\bf \% Best} & {\bf Worst-Case} & {\bf Replicator} & {\bf Group-1} & {\bf Group-2} & {\bf Rank Summary} \\ 
& {\bf Average} & {\bf Score} & {\bf Score} & {\bf Dynamic} & {\bf Tourney} & {\bf Tourney} & {\tiny \bf Best -- Mean -- Worst} \\ 
\hline

{\sc S++} & ~~1, ~~1, ~~1 & ~~2, ~~1, ~~2 & ~~1, ~~1, ~~1 & ~~1, ~~1, ~~1 & ~~1, ~~1, ~~2 & ~~1, ~~1, ~~1 & ~~1 -- ~~1.2 -- ~~2 \\ 
{\sc Manipulator} & ~~3, ~~2, ~~3 & ~~4, ~~3, ~~8 & ~~5, ~~2, ~~4 & ~~6, ~~4, ~~3 & ~~5, ~~3, ~~3 & ~~5, ~~2, ~~2 & ~~2 -- ~~3.7 -- ~~8 \\ 
{\sc Bully} & ~~3, ~~2, ~~1 & ~~3, ~~2, ~~1 & ~~7, 13, 20 & ~~7, ~~3, ~~2 & ~~6, ~~2, ~~1 & ~~6, ~~3, ~~5 & ~~1 -- ~~4.8 -- 20 \\ 
{\sc S++/simple} & ~~5, ~~4, ~~4 & ~~8, ~~5, ~~9 & ~~4, ~~6, 10 & 10, ~~2, ~~6 & ~~8, ~~4, ~~6 & ~~9, ~~4, ~~6 & ~~2 -- ~~6.1 -- 10 \\ 
{\sc S} & ~~5, ~~5, ~~8 & ~~6, ~~7, 10 & ~~3, ~~3, ~~8 & ~~5, ~~5, ~~8 & ~~7, ~~5, ~~9 & ~~7, ~~5, ~~9 & ~~3 -- ~~6.4 -- 10 \\ 
{\sc Fict.~Play} & ~~2, ~~8, 14 & ~~1, ~~6, 10 & ~~2, ~~8, 16 & ~~3, 12, 15 & ~~2, ~~8, 12 & ~~4, ~~9, 14 & ~~1 -- ~~8.1 -- 16 \\ 
{\sc MBRL-1} & ~~6, ~~6, 10 & ~~5, ~~4, ~~7 & ~~8, ~~7, 14 & 11, 11, 13 & ~~9, ~~7, 10 & ~~8, ~~7, 10 & ~~4 -- ~~8.5 -- 14 \\ 
{\sc EEE} & 11, ~~8, ~~7 & 14, ~~9, ~~5 & ~~9, ~~4, ~~2 & 14, 10, ~~9 & 13, ~~9, ~~8 & 13, 10, ~~8 & ~~2 -- ~~9.1 -- 14 \\ 
{\sc MBRL-2} & 14, ~~5, ~~5 & 13, ~~8, ~~6 & 19, ~~5, ~~3 & 18, ~~9, ~~4 & 18, ~~6, ~~5 & 18, ~~6, ~~4 & ~~3 -- ~~9.2 -- 19 \\ 
{\sc Mem-1} & ~~6, ~~9, 13 & ~~7, 10, 21 & ~~6, ~~9, 17 & ~~2, ~~6, 10 & ~~3, 10, 17 & ~~2, ~~8, 15 & ~~2 -- ~~9.5 -- 21 \\ 
{\sc M-Qubed} & 14, 20, ~~4 & 15, 20, ~~3 & 15, 19, ~~5 & 17, 19, ~~5 & 17, 21, ~~4 & 16, 21, ~~3 & ~~3 -- 13.2 -- 21 \\ 
{\sc Mem-2} & ~~9, 11, 20 & ~~9, 11, 22 & 13, 17, 22 & ~~4, 13, 19 & ~~4, 13, 25 & ~~3, 12, 20 & ~~3 -- 13.7 -- 25 \\ 
{\sc Manip-Gf} & 11, 11, 21 & 12, 12, 19 & 12, 11, 19 & ~~9, ~~7, 20 & 12, 14, 20 & 11, 13, 21 & ~~7 -- 14.2 -- 21 \\ 
{\sc WoLF-PHC} & 17, 11, 13 & 18, 14, 14 & 18, 14, 18 & 16, 14, 14 & 16, 11, 11 & 15, 11, 11 & 11 -- 14.2 -- 18 \\ 
{\sc QL} & 17, 17, ~~7 & 19, 19, ~~4 & 17, 18, ~~7 & 19, 18, ~~7 & 19, 20, ~~7 & 19, 18, ~~7 & ~~4 -- 14.4 -- 20 \\ 
{\sc gTFT (Godfather)} & 11, 14, 22 & 11, 15, 20 & 11, 16, 23 & ~~8, ~~8, 22 & 10, 16, 21 & 10, 15, 22 & ~~8 -- 15.3 -- 23 \\ 
%{\sc ~~~(gTFT)} & & & & & & & \\
{\sc EEE/simple} & 20, 15, 11 & 20, 17, 12 & 20, 10, ~~9 & 20, 16, 11 & 24, 15, 14 & 20, 16, 13 & ~~9 -- 15.7 -- 24 \\ 
{\sc Exp3} & 19, 23, 11 & 16, 23, 15 & 16, 23, ~~6 & 15, 23, 12 & 15, 25, 13 & 17, 25, 12 & ~~6 -- 17.2 -- 25 \\ 
{\sc CJAL} & 24, 14, 14 & 25, 14, 13 & 24, 12, 15 & 24, 17, 16 & 20, 12, 16 & 22, 14, 16 & 12 -- 17.3 -- 25 \\ 
{\sc WSLS} & ~~9, 17, 24 & 10, 16, 24 & 10, 20, 24 & 12, 20, 24 & 11, 17, 24 & 12, 17, 25 & ~~9 -- 17.6 -- 25 \\ 
{\sc GIGA-WoLF} & 14, 19, 23 & 17, 18, 23 & 14, 15, 21 & 13, 15, 23 & 14, 18, 22 & 14, 19, 23 & 13 -- 18.1 -- 23 \\ 
{\sc WMA} & 21, 21, 15 & 21, 21, 16 & 22, 21, 12 & 22, 21, 17 & 21, 19, 15 & 23, 20, 17 & 12 -- 19.2 -- 23 \\ 
{\sc Stoch.~FP} & 21, 21, 15 & 22, 22, 17 & 23, 22, 11 & 23, 22, 18 & 25, 24, 18 & 25, 22, 18 & 11 -- 20.5 -- 25 \\ 
{\sc Exp3/simple} & 21, 24, 16 & 23, 24, 18 & 21, 24, 13 & 21, 24, 21 & 22, 22, 19 & 21, 23, 19 & 13 -- 20.9 -- 24 \\ 
{\sc Random} & 24, 25, 25 & 24, 25, 25 & 25, 25, 25 & 25, 25, 25 & 23, 23, 23 & 24, 24, 24 & 23 -- 24.4 -- 25 \\ 
\hline
\multicolumn{8}{c}{~} \\
\end{tabular}}}
%\subfigure[Illustration of S++'s learning dynamics in Chicken]{\includegraphics[height=2.05in]{figs/heatmap2.pdf}}~~~~~~~~~~~~~
%\subfigure[Self play in a Prisoner's Dilemma]{\includegraphics[height=2.05in]{figs/speedSI.pdf}}
\subfigure[Illustration of S++'s learning dynamics in Chicken]{\includegraphics[height=2.05in]{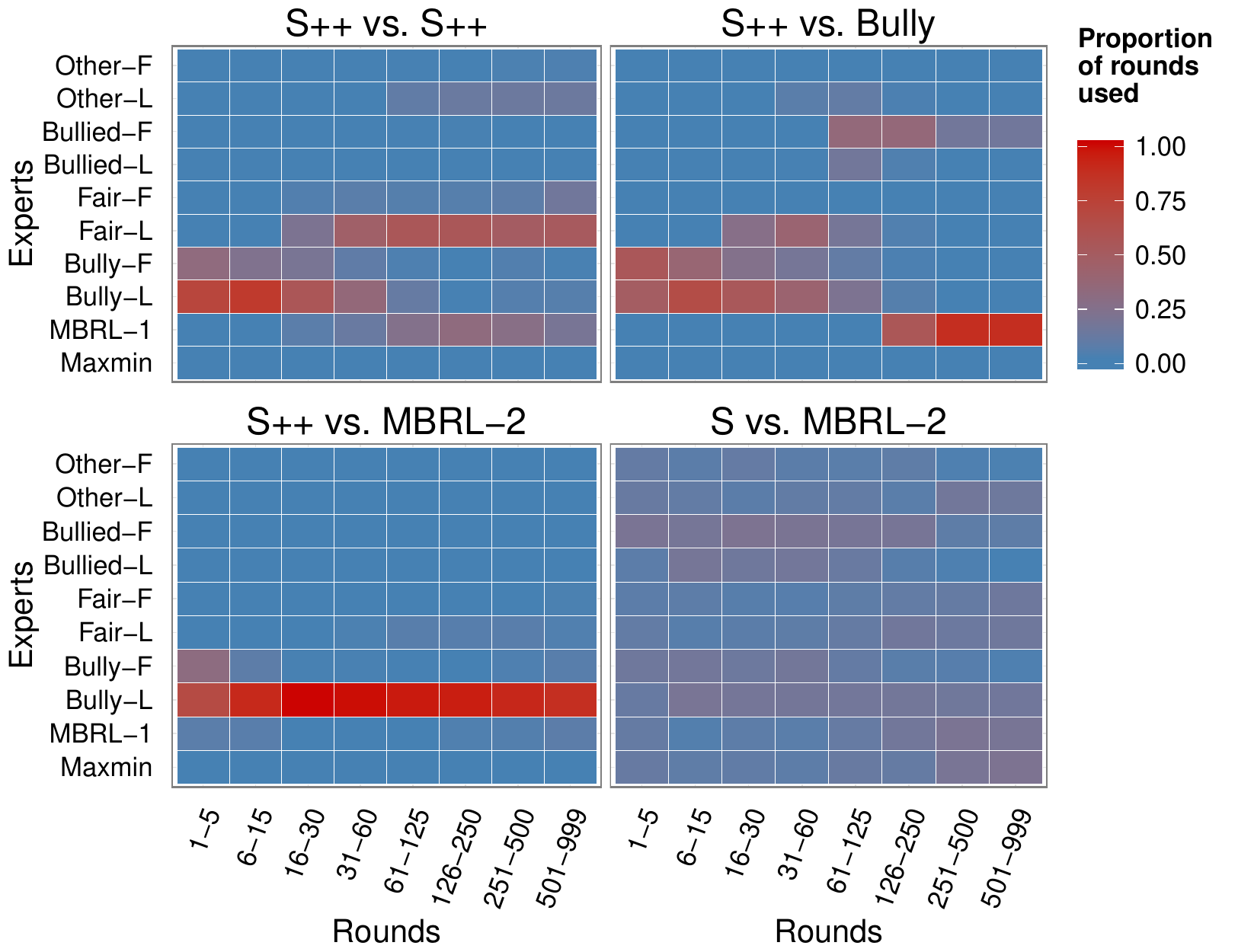}}~~~~~~~~~~~~~
\subfigure[Self play in a Prisoner's Dilemma]{\includegraphics[height=2.05in]{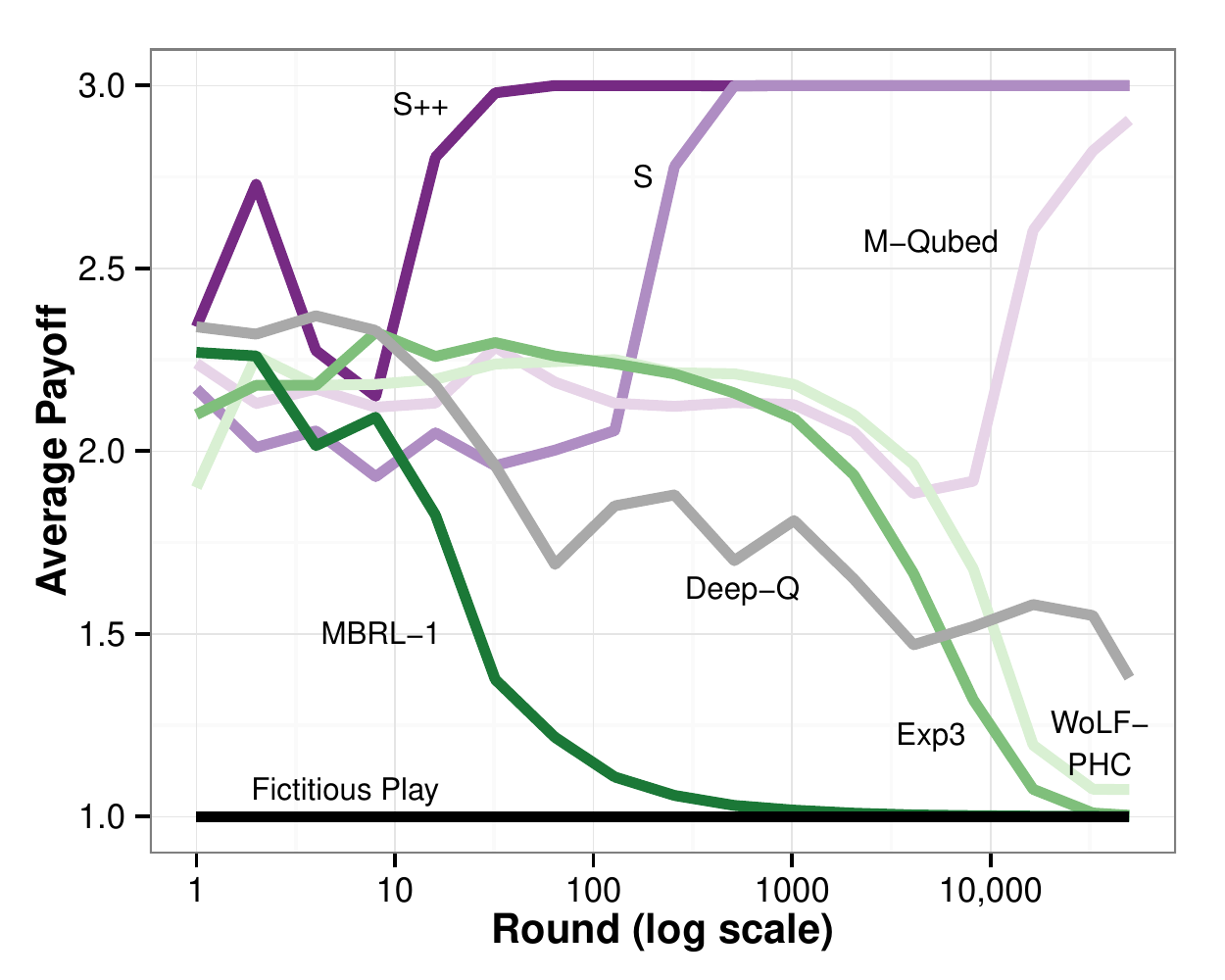}}
\end{center} \vspace{-.1in}
\caption{\small {Selected results comparing the performance of representative algorithms across the periodic table of 2x2 games (Figure~\ref{fig:periodic_table}; see also SI.A.3) when paired with other algorithms.  {\bf (a)}~The rankings of 25 algorithms with respect to six performance metrics (see SI.B.2).  A lower rank indicates higher performance.  For each metric, the algorithms are ranked in 100-round, 1000-round, and 50,000-round games, respectively.  Example: the 3-tuple $3, 2, 1$ indicates the algorithm was ranked $3^{\rm rd}$, $2^{\rm nd}$, and $1^{\rm st}$ in 100, 1000,~and 50,000-round games, respectively.  {\bf (b)}~An illustration of S++'s learning dynamics in Chicken.  For ease of understanding, experts are categorized into groups (see SI.C).  Top-left: When (unknowingly) paired with an agent that uses the same algorithm, S++ initially seeks to bully its associate, but then switches to fair, cooperative experts when attempts to exploit are unsuccessful.  Top-right: When paired with {\sc Bully}, S++ learns the best response, which is to be bullied, achieved by playing MBRL-1, Bully-L, or Bully-F.  Bottom-left: S++ quickly learns to play experts that bully MBRL-2.  Bottom-right: On the other hand, algorithm S does not learn to consistently bully MBRL-2, showing that S++'s pruning rule (Eq.~\ref{eq:theset}) enables it to teach MBRL-2 to accept being bullied, thus producing high payoffs for S++.  These results are averaged over 50 trials each.  {\bf (c)}~The average per-round payoffs of various machine-learning algorithms over time in self play in a traditional (0-1-3-5)-Prisoner's Dilemma in which mutual cooperation produces a payoff of 3 and mutual defection produces a payoff of~1.  Results are the averages of 50 trials.  Among the machine-learning algorithms we evaluated, S++ is unique in its ability to quickly form successful relationships with other algorithms across the set of 2x2 games.}}
\label{fig:perfSummary}
\end{figure}

With the goal of identifying successful algorithmic mechanisms for playing arbitrary repeated games, we selected and evaluated 25 existing algorithms (see Figure~\ref{fig:perfSummary}a) with respect to six different performance metrics (see SI.B.2) across the periodic table of 2x2 games.  These representative algorithms included classical algorithms such as (generalized) generous tit-for-tat (i.e., {\sc Godfather}) and win-stay-lose-shift (WSLS)~\cite{Nowak1993}, evolutionarily evolved memory-one and memory-two stochastic strategies~\cite{Iliopoulous2010}, machine-learning algorithms (including reinforcement learning), belief-based algorithms~\cite{FictitiousPlay}, and expert algorithms~\cite{exextrade,CrandallJAIR2014}.

Results of this evaluation are summarized in Figure~\ref{fig:perfSummary}a.  Detailed analysis is provided in SI.B.  We make two high-level observations.  First, it is interesting to observe which algorithms were less successful in these evaluations.  For instance, while generalized tit-for-tat, WSLS, and memory-one and memory-two stochastic strategies (e.g., {\sc Mem-1} and {\sc Mem-2}) are successful in prisoner's dilemmas, they are not consistently effective across the full set of 2x2 games.  These algorithms are particularly ineffective in longer interactions, as they do not effectively adapt to their associate's behavior.  Additionally, algorithms that minimize regret (e.g., Exp3~\cite{auer95gambling}, GIGA-WoLF~\cite{GIGA-WoLF}, and WMA~\cite{FreundSchapire95}), which is the central component of world-champion computer poker algorithms~\cite{bowling2015heads}, also performed poorly.

Second, while many algorithms had high performance with respect to some measure, only S++~\cite{CrandallJAIR2014} was a top-performing algorithm across all metrics at all game lengths.  Furthermore, it maintained this high performance in each class of game and when associating with each class of algorithm (see SI.B.5). S++ learns to cooperate with like-minded associates, exploit weaker competition, and bound its worst-case performance (Figure~\ref{fig:perfSummary}b).  Perhaps most importantly, whereas many machine-learning algorithms do not learn cooperative behavior until after thousands of rounds of interaction (if at all), S++ tends to do so within relatively few rounds of interaction (Figure~\ref{fig:perfSummary}c), likely fast enough to support interactions with people.

\subsection{S\#: A Machine-Learning Algorithm that Talks}

S\# is derived from S++~\cite{CrandallJAIR2014}, an expert algorithm that combines and builds on decades of research in computer science, economics, and the behavioral and social sciences.  S++ uses the description of the game environment to compute a diverse set of experts, each of which uses distinct mathematics and assumptions to produce a strategy over the entire space of the game.  S++ then uses a meta-level control strategy based on aspiration learning~\cite{Simon1956,Karandikar,StimpsonIJCAI} to dynamically reduce this set of experts.  Formally, let $E$ denote the set of experts computed by S++.  In each epoch (beginning in round~$t$), S++ computes the potential $\rho_j(t)$ of each expert $e_j \in E$, and compares this potential with its aspiration level $\alpha(t)$ to form a reduced set $E(t)$ of experts:
\begin{eqnarray}
E(t) = \{e_j \in E : \rho_j(t) \geq \alpha(t) \}.
\label{eq:theset}
\end{eqnarray}
This reduced set consists of the experts that S++ believes could potentially produce satisfactory payoffs.  It then selects one expert $e_{\rm sel}(t) \in E(t)$ using a satisficing decision rule~\cite{Karandikar,StimpsonIJCAI}.  Over the next $m$ rounds, it follows the strategy prescribed by $e_{\rm sel}(t)$, after which it updates its aspiration level as follows:
\begin{eqnarray}
\alpha(t+m) \gets \lambda^m \alpha(t) + (1 - \lambda^{m}) R,
\label{eq:spp_aspirationupdate}
\end{eqnarray}
where $\lambda \in (0,1)$ is the learning rate and $R$ is the average payoff obtained by S++ in the last $m$ rounds.  It also updates each expert $e_j \in E$ based on its peculiar reasoning mechanism, and then begins a new epoch.

These results demonstrate the ability of S++ to effectively establish and maintain profitable long-term relationships with machines in arbitrary repeated games.  Does S++ also learn to form cooperative relationships with people?

S\# differs from S++ in two ways.  First, the partner's proposed plans, signaled via speech acts, are used to further reduce the set of experts that S\# considers selecting (Figure~\ref{fig:Ssharp}c).  Formally, let $E_{\rm cong}(t)$ denote the set of experts in round $t$ that are \emph{congruent} with the last joint plan proposed by S\#'s partner (see SI.C.2.2).  Then, S\# considers selecting experts from the following set:
\begin{eqnarray}
E(t) = \{e_j \in E_{\rm cong}(t) : \rho_j(t) \geq \alpha(t) \}.
\label{eq:theset2}
\end{eqnarray}
If this set is empty (i.e., no desirable options are congruent with the partner's proposal), $E(t)$ is calculated as in the original S++ (Eq.~\ref{eq:theset}).  Second, S\# also extends S++ by generating speech acts that convey the ``stream of consciousness'' of the algorithm (Figure~\ref{fig:Ssharp}d).  Specifically, S\# generates a finite-state machine with output for each expert.  Given the state of the expert and the game outcomes, the state machine of the currently selected expert produces speech derived from a predetermined set of phrases.  The set of speech acts, which are largely game-generic (though some adaptations must be made for multi-stage games; see SI.E.3.4) allows S\# to provide feedback to its partner, make threats, provide various explanations to manage the relationship, and propose and agree to plans.

See SI.C for an in-depth description of S\#.

%{\color{red} Add more details; note that the algorithm is described in the SI.C in detail.}

\bibliography{biblio}
\bibliographystyle{unsrt}

\newpage

The supplementary information (SI) for this paper is available {\color{blue} \href{https://www.dropbox.com/s/o7x2wbig3o8jlzq/CoopMachines_SI.pdf?dl=0}{here}}.

\end{document}